\documentclass[
]{ceurart}

\usepackage{listings}
\usepackage{amsmath}
\usepackage{amsthm}
\newtheorem{proposition}{Proposition}
\begin{document}

\copyrightyear{2022}
\copyrightclause{Copyright for this paper by its authors.
  Use permitted under Creative Commons License Attribution 4.0
  International (CC BY 4.0).}

\conference{NeSy 22: 16th International Workshop on
Neural-Symbolic Learning and Reasoning}

\title{Differentiable Rule Induction with Learned Relational Features}

\author[1,2]{Remy Kusters}[%
email=remy.kusters@ibm.com]
\cormark[1]
\address[1]{IBM Research, Orsay, France}
\address[2]{IBM  France Lab, Orsay, France}
\address[3]{Inria Saclay Ile-de-France, Palaiseau, France}

\author[1,2]{Yusik Kim}[]
\author[2,3]{Marine Collery}[]
\author[2]{Christian {de Sainte Marie}}[]
\author[1,2]{Shubham Gupta}[]

\cortext[1]{Corresponding author.}

\begin{abstract}
  Rule-based decision models are attractive due to their interpretability. However, existing rule induction methods often result in long and consequently less interpretable rule models. This problem can often be attributed to the lack of appropriately expressive vocabulary, i.e., relevant predicates used as literals in the decision model. Most existing rule induction algorithms presume pre-defined literals, naturally decoupling the definition of the literals from the rule learning phase. In contrast, we propose the Relational Rule Network (R2N), a neural architecture that learns literals that represent a linear relationship among numerical input features along with the rules that use them. This approach opens the door to increasing the expressiveness of induced decision models by coupling literal learning directly with rule learning in an end-to-end differentiable fashion. On benchmark tasks, we show that these learned literals are simple enough to retain interpretability, yet improve prediction accuracy and provide sets of rules that are more concise compared to state-of-the-art rule induction algorithms.
\end{abstract}


\maketitle

\section{Introduction}

Over the last decade, black box decision models (e.g. neural networks) are increasingly used in high stakes decision making, yet there has been equally growing concern over their use \citep{ProPublica}. Regulations in various industries are demanding accountability and transparency from the models involved in the decision making process. Besides the regulatory concerns, the end-users of many complex decision models are also increasingly demanding interpretability of the final decision. Rule based models are a potential solution but, to date, most rule-based decision support systems do not support learning practically useful decision models \emph{directly from the data}, often due to the resulting rules being too long to be considered interpretable \citep{sainte2021,ijcai2020-678}. Focusing on learning rule models which model higher-level concepts expressed in terms of the lower-level input features, rather than rule models that \textit{only} use these lower-level input features has been brought forward as one of the primary avenues to improve the expressiveness and interpretability of rule models \citep{sainte2021, dumanvcic2017demystifying}.
The scope of this paper is learning rule models, expressed in Disjunctive Normal Form (DNF), for classification tasks of tabular data. Of particular interest is to enrich the representation language used by rule models with higher-level latent representations which are functions of lower-level input features. This is in contrast to most state-of-the-art rule learning algorithms which use pre-defined literals (either hand-crafted or obtained from a-priori binarized features) as literals in the rule model (e.g., BRCG \citep{dash2018boolean}, RIPPER \citep{cohen1995fast}, CORELS \citep{JMLR:v18:17-716}). The main difficulty of learning the appropriate literals is in incorporating feedback from the rule model that uses them. One way to achieve this is through learning the literals and the rule model that uses them \emph{jointly} via a single differentiable neural architecture.

\begin{figure}[t]
    \centering
    \includegraphics[width=0.5\textwidth]{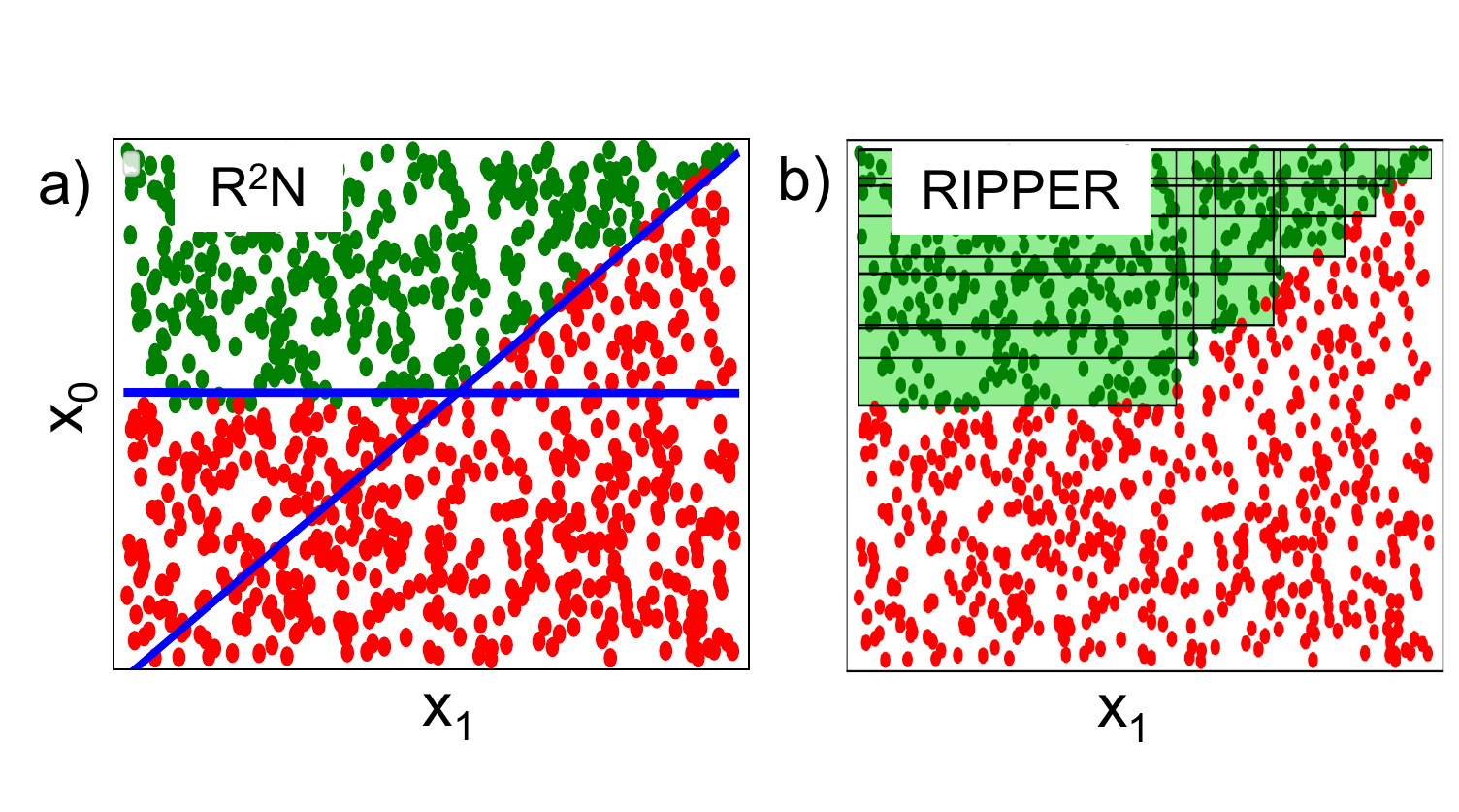}
        \caption{Decision boundaries for Eq. (\ref{eq:ex}) (a) R2N, and (b) the RIPPER algorithm \cite{cohen1995fast}. The decision boundaries learned by R2N (a) correspond to $x_0 > 0.5$ and $x_1/x_0 > 1.0$, while the RIPPER algorithm (b) learns nine rules that correspond to the nine rectangles inside the green area.}
        \label{fig:2}
\end{figure}

We propose the \emph{Relational Rule Network (R2N)}: a three-layer neural network which \textit{learns literals} in the first layer, each representing a partition of the input feature space delimited by a hyperplane, which we call a \emph{halfspace}. The second and third layers map the binary vector of literals to a binary predictions, encoding a crisp logical formula in the form of a DNF. We consider halfspaces as literals as they are simple enough to retain interpretability, yet significantly improve expressivity and model accuracy. Encoding this structure in an end-to-end differentiable neural network has the benefit that the learning of the literals is directly informed by their usage in the rule. To showcase the value of using appropriate predicates as literals in a rule model, we present a simple toy example: Given a dataset with two numerical input features $(x_0,x_1)$ and a corresponding label $y$ determined by the ground truth
\begin{align}
\label{eq:ex}
    &\text{if} \left( x_0/x_1 > 1.0  \land  x_0 > 0.5 \right)  \text{ then }  \text{class = True;} \text{ else} \text{ class = False;}
\end{align}
the task is to find a rule model that accurately predicts the label given the input.
This model involves two distinct decision boundaries: $x_0 = 0.5$ and $x_0/x_1 = 1.0$. Training our model with default parameters (see Appendix \ref{appendix:loss}) recovers the ground truth (see Fig. \ref{fig:2}) whereas when literals are predefined as univariate value comparisons (e.g. $x_0 < 0.2$), we learn a longer DNF. This is because each conjunction represents a rectangle in the feature space $(x_0,x_1)$, and many of them are needed to adequately approximate the sloped decision boundary, $x_0/x_1 = 1$ (Fig. \ref{fig:2}b). Therefore, by improving the expressivity of the rule model with the ability to represent ratio as a single literal rather than a conjunction of many, we simplified the model and improved accuracy. 

We summarize our contribution as follows: i) We propose a neural network architecture (R2N) that learns a DNF together with the literals it uses. The literals represent halfspaces, whose boundary hyperplanes are \emph{learned} by the network. ii) We demonstrate that compared to SOTA algorithms (RIPPER, BRCG, DR-Net, etc.) we improve sparsity and interpretability of the learned rule model both for a set of benchmark problems where the underlying decision models are known as well as where they are not.

\section{Related works}

Our work contributes to the large body of work on neuro-symbolic rule learning systems that represent atoms and clauses as individual nodes in a neural network (NN) and that learn rules as configurations of weights on the edges that link these nodes. This approach is different from existing \textit{distillation}-based approaches that construct a simpler interpretable network to mimic the behavior of a complex black-box model \citep{HintonEtAl:2015:DistillingTheKnowledgeInANeuralNetwork,zilke2016}. While they generate post-hoc explanations that only approximate a model, R2N and the like natively learn explainable rules: the rules do not approximate the NN, they are equivalent to it by construction.

The representation of rules in R2N is essentially the same as in C-IL2P \cite{avila1999} and followers (e.g. \cite{francca2014,gao2022learning,marra2020relational}). R2N differs from C-IL2P and similar systems in two main ways: whereas they focus on approximating logic programs and their semantics in discrete domains (although infinite and continuous domains can also be dealt with, it remains in a probabilistic setting \cite{belle2020symbolic}), R2N learns crisp classification rules in continuous numerical domains, like \cite{qiao2020learning,beck2021investigation}. However, our approach differs from \cite{qiao2020learning} in two regards: i) the encoding and ii) the binarization of the rules as we discuss in Section 4. 

In addition, R2N learns a set of literals that map the $n$-dimensional continuous numerical input space to an $m$-dimensional Boolean space in which the rule models are learned. Similar neuro-symbolic systems require binarized input data using predefined predicates \cite{avila1999,qiao2020learning, beck2021investigation, Fu1994}; are limited to learning univariate threshold values \cite{wang2021scalable}; or learn latent relations in a finite Herbrand base \cite{dumanvcic2017demystifying, marra2020relational,sourek2015}. The latter systems rely on prior knowledge about the relations to be learned \cite{dumanvcic2017demystifying,sourek2015}, whereas R2N does not require prior knowledge. Rather than predicate invention, learning literals in R2N can be considered as a form of propositionalization \cite{kramer2000bottom}. Contrary to other systems relying on it \cite{gao2022learning, kramer2000bottom, krogel2003comparative}, propositionalization is intrinsically part of the rule learning in R2N: indeed, R2N is able to learn a more useful grounding of the learned literals in the input space without prior knowledge because the literals are learned as part of the supervised rule learning process. This is similar to the way DeepProbLog uses the semantics of the program to supervise the grounding of predicates \cite{manhaeve2018deepproblog}.

Extending the branching conditions from simple value comparisons to linear 
halfspaces has been proposed for decision trees \citep{NEURIPS2020_1373b284}. 
While it is possible to translate the resulting decision tree to a DNF, such mapping usually results in excessively long rules, hindering interpretability.  Furthermore, single literal learning layer that grounds the learned literals using hyperplanes in R2N can be replaced with an arbitrarily complex NN (learning hypersurfaces) at the cost of interpretability. 

\section{Rule language} 

The rule model that we consider for binary classification is a DNF, whose formal grammar is defined by the following production rules, 
\begin{align*}
    \tt{rule model} &\rightarrow \text{if } \tt{ DNF } \text{ then }  \text{ class = True; } \text{else} \text{ class = False} \\
    \tt{DNF} &\rightarrow \tt{conjunction }\vert \tt{ conjunction }\lor\tt{ DNF}\\
    \tt{conjunction} &\rightarrow\tt{literal } \vert \tt{ literal }\land \tt{ conjunction}\\
    \tt{literal} &\rightarrow \text{\underline{$\phi_{1}$}$\vert$ ... $\vert$ \underline{$\phi_{m}$} $\vert$ \underline{$\phi_{m+1}$}$\vert$ ... $\vert$ \underline{$\phi_{M}$}}
\end{align*}
The \underline{underlined} terms are terminal symbols, and the {\tt typewriter} terms are the non-terminal symbols. Among the predicates (of arbitrary arity) in our dictionary of literals, some have learnable parameters, \{\underline{$\phi_1$},...,\underline{$\phi_m$}\}, e.g., coefficients of a hyperplane defining a halfspace, and some do not, \{\underline{$\phi_{m+1}$},...,\underline{$\phi_M$}\}, meaning they are pre-determined (see Fig. \ref{fig:1}). We call the subset of literals with learnable parameters \emph{learned literals}, and those without learnable parameters \emph{predefined literals}. Note that negations are not part of the rule language. Therefore, negated predicates, if desired, must be explicitly included in the dictionary of literals.
\section{The Relational Rule Network (R2N)}

\begin{figure}[t]
    \centering
    \includegraphics[width=0.5\textwidth]{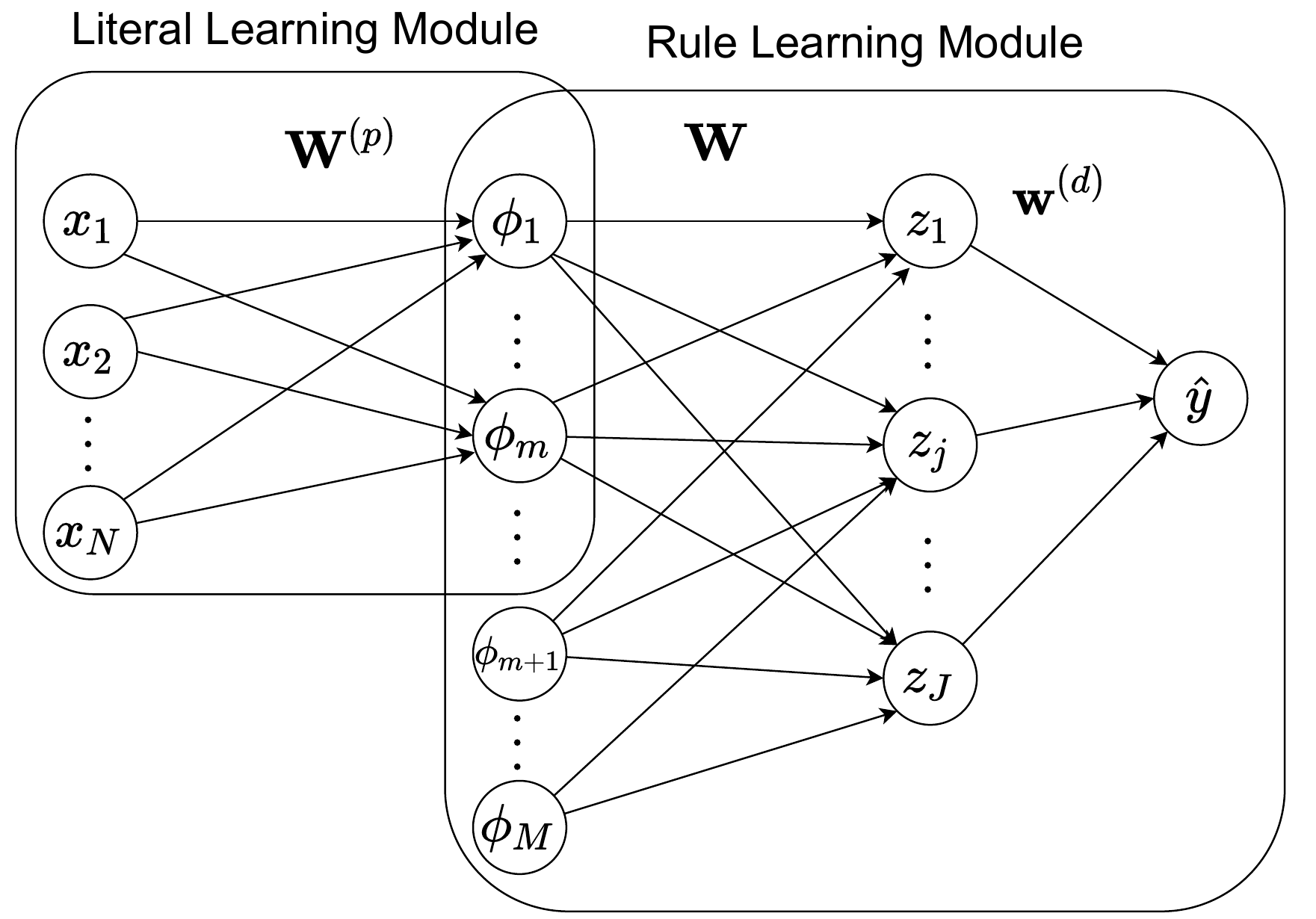}
    \caption{Relational Rule Network (R2N): The literal learning module encodes halfspaces and evaluates to Boolean literals $\phi$. Alongside these learned literals, zero or more predefined literals $\phi_{m+1},..., \phi_M$ are supplied directly to the rule learning module. The rule learning module uses these literals in the conjunctions $z$ in the first layer. In the final layer, the relevant conjunctions are selected into a DNF. }
    \label{fig:1}
\end{figure}

The Relational Rule Network (R2N) is composed of two modules connected sequentially: 
\begin{enumerate}
    \item The {\bf literal learning module} that learns halfspaces to be used as literals of the rule language.
    \item The {\bf rule learning module} that maps the binary vector of evaluated literals to a single binary prediction of the class label, in such a way that it encodes an equivalent logical formula in disjunctive normal form (DNF). 
\end{enumerate}
 Any predefined literals we want to use, including categorical value comparisons, are directly supplied to the input layer of the rule learning module. See Fig. \ref{fig:1} for an illustration. 


We introduce the following notation for the rest of this paper:
\begin{itemize}
    \item $x_i$: Numerical input variable $i$.
    \item $\phi_i({\mathbf x})$: The  $i$-th literal in our rule language. Learned literals ($i=1,...,m$) represent halfspaces in the N-dimensional space of $
    \mathbf{x}$. We omit the argument when the context is clear.
    \item $z_j$: The $j$-th conjunction formed by a combination of literals. 
\end{itemize}
Unless stated otherwise, normal-faced lower-case variables with subscripts (e.g., $x_i$) represent scalar components of its bold-faced counterparts, representing vectors (lower-case; $\mathbf{x}$) or matrices (upper-case; $\mathbf{X}$).
All vectors are column vectors.

\subsection{The literal learning module}

To enrich the vocabulary of our rule language, we consider a neural network layer that learns predicates representing halfspaces of the numerical feature space.
These learned predicates can then be used as literals in the rule language.
The literal learning module is represented by a single layer perceptron with a learnable bias $b_i$ and weight vector $\mathbf{w}^{(p)}_j$ for each halfspace $\phi_i$ that we wish to learn (See Fig. \ref{fig:1}). Extending this to multiple output nodes using matrix notation, the transformation is represented by
\begin{equation}
\boldsymbol\phi(\mathbf{x})= \sigma\left(\frac{\mathbf{x}^T\mathbf{W}^{(p)} + \mathbf{b}}{\tau}\right),
\end{equation}
where $\mathbf{W}^{(p)}$ is the weight matrix, $\mathbf{b}$ the corresponding biases, $\sigma$ is the sigmoid activation function applied component-wise, and $\tau$ is a temperature parameter that changes according to a cooling schedule. Observing that $\sigma(x/\tau)$ converges\footnote{It pointwise converges on $\mathbb{R}\setminus\{0\}$.} to the Heaviside step function as $\tau\rightarrow 0$, we effectively approach strict binarization at the end of the cooling schedule (step function is used during testing), while avoiding zero gradients during training.

\subsection{The rule learning module}

The rule learning module maps the binary vector of literals to a single binary prediction of the class label, in such a way that it encodes an equivalent logical formula in DNF. We specify this rule learning module as a two-layer neural network architecture respectively implementing an \textbf{AND} and a subsequent \textbf{OR} operation, similar to \cite{avila1999,qiao2020learning}.

\paragraph{AND-layer} 
The \textbf{AND}-layer maps a binary vector of literals ${\boldsymbol\phi} = (\phi_1,...,\phi_M)$ to another binary vector $(z_1,...,z_J)$ whose components represent conjunctions made from some combination of the input literals.
This combination is specified by a binary vector ${\mathbf w}_j$ indicating which of the $M$ literals are present in conjunction $z_j$. The $i$-th component $w_{ij}$ of ${\mathbf w}_j$ is 1 if literal $\phi_i$ is included in conjunction $z_j$ and 0 otherwise \citep{beck2021investigation}. 
\begin{proposition}
\label{prop:and}
Let $w_{ij}$ be binary parameter and $\phi_i$ be Boolean variables whose values can be represented by binary values. Then the mappings
\begin{align}
\label{eq:and_transform}
     \boldsymbol\phi\mapsto\bigwedge_{i:w_{ij}=1}\phi_i \;\;\;\text{   and   }  \;\;\; \boldsymbol\phi\mapsto 1-\min\left\{\sum_i w_{ij}(1-\phi_i), 1\right\}
\end{align}
are equivalent. Proof: See Appendix  \ref{app:proof_and}
\end{proposition}

The \textbf{AND}-layer implements transformation (\ref{eq:and_transform}) as a computational graph.
Although the parameter we want to eventually learn is the binary weight matrix ${\mathbf W}$, we learn it indirectly through  reparameterizing it with a scaled sigmoid function that converges to a Heaviside step function in the limit, similarly as in the literal learning module. The learnable parameter $u_{ij}$ for each binary weight $w_{ij}$ is continuous and unconstrained:
\begin{align}
\label{eq:and_scaled_sigmoid}
    w_{ij} = \sigma(u_{ij}/\tau).
\end{align}
The resulting computational graph defined by equations (\ref{eq:and_transform}) and (\ref{eq:and_scaled_sigmoid}) is differentiable with non-zero derivatives. Note that ensuring non-zero gradients during the backwards pass, rather than using a straight through estimator, as in \cite{qiao2020learning}, is essential to ensure robust training, as described in \cite{bengio2013estimating}. 

\paragraph{OR-layer} 
The \textbf{OR}-layer maps a binary vector $\mathbf{z}$ to a single binary value by implementing a logical disjunction operation among select components of $\mathbf{z}$.
The binary weights ${\mathbf w^{(d)}}$ of this layer encode which of the conjunctions learned in the \textbf{AND}-layer are to be included in the final disjunction.
We seek a differentiable approximation that approaches the non-differentiable \textbf{OR} operation in the limit, similar to what we have done for the \textbf{AND}-layer.
\begin{proposition}
\label{prop:or}
Let $w_{j}^{(d)}$ be binary parameters and $z_j$ be Boolean variables. Then the mappings
\begin{align}
\label{eq:or_transform}
     \mathbf{z}\mapsto\bigvee_{j:w_j^{(d)}=1}z_j \;\;\;\text{   and   }\;\;\;
     \mathbf{z}\mapsto\max\left\{w^{(d)}_j z_j; j=1,...,J\right\} 
\end{align}
are equivalent. Proof: See Appendix \ref{app:proof_or}
\end{proposition}

The binary weight vector ${\mathbf w^{(d)}}$ is learned exactly the same way as we learn the weight matrix $\mathbf{W}$ of the \textbf{AND}-layer, through reparameterizing the binary weights using a scaled sigmoid:
\begin{align}
\label{eq:or_scaled_sigmoid}
    w_j^{(d)}=\sigma(v_j/\tau).
\end{align}
The \textbf{OR}-layer implements the computational graph defined by equations (\ref{eq:or_transform}) and (\ref{eq:or_scaled_sigmoid}), where $v_j$ are the continuous and free learnable parameters.

\paragraph{Decoding the DNF from the trained network}
The 2-layer network composing the \textbf{AND} and \textbf{OR} layers can be represented as a composition of the two differentiable functions (\ref{eq:and_transform}) and (\ref{eq:or_transform}).
As a direct result of Propositions \ref{prop:and} and \ref{prop:or}, this network implements the evaluation of the DNF
\begin{align}
    \boldsymbol\phi\mapsto\bigvee_{j:w_j^{(d)}=1}\bigwedge_{i:w_{ij}=1}\phi_i.
\end{align}
The binary weights $w_{ij}$ and $w_j^{(d)}$ can be retrieved by inspecting the sign of the learned parameters $u_{ij}$ and $v_j$.

 Note, however, that during training the network weights are not binary when the temperature is different from 0. Hence proposition 1 is not strictly applicable at training time, so there is no guarantee that at this stage the network output and the output of the decoded DNF are equivalent. However, we can make the outputs arbitrarily close to each other by driving the temperature down to zero. We observed that the difference between the network output and the output of the decoded DNF is negligible for reasonably cold temperatures ($\tau = 10^{-4}$). Therefore at the end of training we learn a crisp DNF as classifcation rule. 

\section{Results}

\begin{table*}[h!]
\begin{center}
\scriptsize
\resizebox{\textwidth}{!}{
\begin{tabular}{|m{4cm}|>{\centering\arraybackslash}m{0.7cm}|>{\centering\arraybackslash}m{0.7cm}|>{\centering\arraybackslash}m{0.7cm}|>{\centering\arraybackslash}m{0.75cm}|>{\centering\arraybackslash}m{0.7cm}|>{\centering\arraybackslash}m{0.75cm}|>{\centering\arraybackslash}m{0.7cm}|>{\centering\arraybackslash}m{0.75cm}|>{\centering\arraybackslash}m{0.7cm}|}
\hline
\multicolumn{10}{|c|}{Results with known ground truth} \\
\hline
\multicolumn{1}{|c|}{Examples} & \multicolumn{2}{c|}{R2N} & \multicolumn{2}{c|}{DR-Net}& \multicolumn{2}{c|}{RIPPER} & \multicolumn{2}{c|}{BRCG}& \multicolumn{1}{c|}{MLP}\\
\cline{2-10}
\multicolumn{1}{|c|}{TRUE IF } & Acc. & $n_r,l_r$ & Acc. & $n_r,l_r$ & Acc. & $n_r,l_r$ & Acc. & $n_r,l_r$& Acc \\
\hline
1) $(x_0 > 0.25) \lor (x_1 < 0.5)$   &1.0&2, 1.0&0.99&28, 1.6&0.99&5, 2.2&1.0&2, 1.0& 1.0 \\
\hline
2) $(x_0/x_1 > 0.5) \lor (x_4 < 0.25)$ & 1.0 & 2, 1.5&0.99&39, 2.1&0.99&22, 2.8&0.98&6, 1.5& 1.0 \\
\hline
3) $(x_0 + 0.5x_1 > 0.5) \lor (x_4 < 0.25)$   &0.99 & 3, 1.7&0.98 &40, 2.3&0.99&15, 2.7&0.97&6, 1.7& 1.0\\
\hline
4) $(x_0 < 0.2 \land x_1 + x_2 > 0.5) \lor $  \newline  $(x_4/x_3 > 1 \land x_1 < 0.5$) &0.99  &4, 2.3 &0.98&  40, 3.7&0.98&23, 4.8&0.94&5, 2.2& 1.0   \\
\hline
5) $(x_4 < 0.2 \land x_0/x_1 > 0.5) \lor $ \newline $ (0.5x_3+0.2 x_1 > 0.5) \lor (x_0 < 0.2)$  &0.99  &3, 1.3 &0.98 &39, 2.0&0.98&15, 3.5&0.97&6, 1.8& 0.99   \\
\hline
\multicolumn{10}{|c|}{Results with unknown ground truth} \\
\hline
\multicolumn{1}{|c|}{Examples} & \multicolumn{2}{c|}{R2N} & \multicolumn{2}{c|}{DR-Net$^{*}$}& \multicolumn{2}{c|}{RIPPER$^{*}$} & \multicolumn{2}{c|}{BRCG$^{*}$}& \multicolumn{1}{c|}{MLP$^{*}$}\\
\cline{2-10}
\hline
Magic &0.86 &6, 3.3 &0.84& 18, 5.2 &0.82& 27,  6.0 &0.84& 22, 3.7 &0.87 \\
\hline
Heloc & 0.72 & 3, 2.0 &0.70&2, 6.3&0.70&12, 5.2&0.69& 1, 1.9 &0.71 \\
\hline
Adult &0.83 & 2, 3.5 &0.83&6, 13.5&0.82&20, 4.7 &0.83&24, 3.8&0.84 \\
\hline
House &0.86  & 5, 2.2 &0.86 &12, 6.3&0.82&41, 7.0 &0.84&5, 5.2 &0.89   \\
\hline

\end{tabular}}
\end{center}
\caption{Accuracy, number of conjunctions in the DNF ($n_r$) and average number of literals per conjunction ($l_r$) for (top) the examples with known ground truth and (bottom) four datasets taken from the UCI ML database. Note that for the columns with an $^*$, the benchmark accuracy and ($n_r, l_r$) are taken from \cite{qiao2020learning} to ensure a fair comparison.}
\label{tab:1}
\end{table*}

In this section we study the accuracy and rule complexity for a set of benchmark problems (i) where the underlying DNF is known and (ii) where this is not the case. We compare with three SOTA methods: RIPPER \citep{cohen1995fast}, DR-Net \citep{qiao2020learning} and BRCG \citep{dash2018boolean}. The hyperparameters, training routine and benchmarks are discussed in the Appendix \ref{appendix:baselines}.

We first consider five simple DNFs with increasing complexity (see Table \ref{tab:1}) and compare the prediction accuracy, number of conjunctions in the DNF ($n_r$) and average number of literals per conjunction ($l_r$) of the R2N with SOTA methods. We generate a dataset of $10^4$ samples from the underlying rules where the five input features ($x_0, ... , x_4)$ are randomly sampled between $[0,1]$. For all examples, R2N obtains near perfect accuracy ($>99\%$) and the DNF we obtain approximate the ground truth very well. E.g., for the most complex example, Ex. 5, we obtain the DNF
$(0.2 x_1 + 0.5 x_3 > 0.5) \lor (x_0 <0.2) \lor (x_0/x_1 > 0.59\ \land\ x_4 < 0.2)$. This example shows that R2N has not only obtained a compact rule model, but has also learned the ground truth literals involving the linear relationships and ratios of input features. The DNF we obtain for the other examples are detailed in Appendix \ref{appendix:rulesets}. 

The three benchmark methods (RIPPER, DR-Net and BRCG) all require binarizing the numerical input features, hence trading off accuracy and length of the optimal DNF (fewer pre-defined literals will result in more compact rule models, but cruder approximations of the decision boundary). We selected the default parameters to make a fair comparisons in this case (See Appendix \ref{appendix:baselines} for a detailed description), but these methods inherently require this accuracy/DNF complexity trade-off. RIPPER and DR-Net provide a comparable accuracy, yet provide DNFs that are considerably longer than R2N while BRCG provides more compact DNFs, but compromises significantly on the prediction accuracy. Note that this discrepancy is particularly important when the decision boundaries cannot be captured by univariate value comparisons as pre-defined literals (Ex. 2-5 in Table \ref{tab:1}). In all cases, the rule language is not expressive enough to describe the linear decision boundary accurately with a sparse set of univariate literals. 

Next we consider four datasets for which the underlying rule models are unknown (Taken from the UCI ML-repository, \cite{Dua:2019}). Note that besides the numerical attributes, these datasets contain categorical data which are added as additional input to the rule learning module using one-hot encoding as described in Section 4. Similarly as for the rule models with known underlying DNF, we obtain significantly shorter DNFs compared to all the SOTA models (see Table \ref{tab:1}). This shows that even though we do not have a priori information of the presence of multi-variate literals, R2N provides more concise rulesets while outperforming SOTA methods on prediction accuracy. The accuracy we obtain with R2N approaches that of a Multi Layer Perceptron\footnote{Comparisons are taken directly from \cite{qiao2020learning}}.  

In Appendix \ref{app:noisesample} and \ref{appendix:sparsity} we show the sensitivity of our approach with respect to the size of the dataset/noise level and sparsity parameters. Overall we show that the obtained rule model and its near-optimal accuracy can still be obtained for noise-levels (randomly flipping a fraction of the labels in the dataset) up to $10\%$ and for training set sizes of the order of $10^3$ training samples.

\section{Outlook}

The method presented in this paper extends the vocabulary of rule-learning algorithms by learning literals that represent halfspaces. R2N can also serve in other applications as part of the domain vocabulary (or ontology); for instance as input predicates for other rule models like RIPPER or BRCG. We show that using these \emph{learned} literals from R2N as input for RIPPER reduces the number of rules that it learns as compared to RIPPER with univariate value comparisons for all examples listed in Table \ref{tab:1} (see Appendix \ref{app:interop}). Note that while in general rules with fewer conjunctions and literals per conjunction can be considered more easily interpretable, true interpretability might have to be examined by subject matter experts.

Although learning literals defined through hyperplanes has been shown to be a good first step towards improving the expressiveness of the rules while retaining interpretability, they are by no means sufficient for all practical applications. The framework we present can be extended to non-linear decision boundaries. A simple yet powerful example of this involves black-box function approximators (e.g. MLPs and LSTMs) to learn non-linear literals using our general machinery from Section 4 for retaining differentiability. While this results in literals that are not explicitly interpretable, the final DNF that use these literals can still be understood and audited by humans who can then try to understand the learned literals in light of the rules that use them. As a concrete example, consider an application where the input data is sequential and the ground-truth class assignments depend on complex aggregates such as variance (e.g., $var(c) > \theta_1 \Rightarrow True$) or counts (e.g. $count(x_t > \theta_2) \leq \theta_3 \Rightarrow False$). Halfspace predicates are inadequate for representing these relationships. However LSTMs can learn binary ``non-linear'' predicates from the data, that can then be used as literals in the rule model. Post-hoc analysis of these literals can then shed light on their meaning, reducing the complex problem of interpreting the entire model to a simpler problem of interpreting individual literals in the context of an a priori understandable rule model.

\begin{acknowledgments}
This work has been partially funded by the French government as part of project PSPC AIDA 2019-PSPC-09, in the framework of the ``Programme d'Investissement d'Avenir''.
\end{acknowledgments}

\bibliography{sample-ceur}

\appendix

\section{Proof of Proposition \ref{prop:and}}
\label{app:proof_and}
As the functions are binary-valued, it suffices to show that one function evaluating to 1 is equivalent to the other function evaluating to 1.
\begin{align}
\nonumber
    \bigwedge_{i:w_{ij}=1}\phi_i=1 &\iff \forall i\;(w_{ij}=1)\rightarrow (\phi_i=1)\\ 
    \nonumber
    &\iff \forall i\;(w_{ij}=0)\lor (\phi_i=1)\\
    \nonumber
    &\iff \forall i\;w_{ij}(1-\phi_i)=0\\
    \label{eq:altout}
    &\iff \sum_i w_{ij}(1-\phi_i)=0\\
    \label{eq:andoutput}
    &\iff 1-\min\left\{\sum_i w_{ij}(1-\phi_i), 1\right\} = 1
\end{align}

\section{Proof of Proposition \ref{prop:or}}
\label{app:proof_or}
\begin{align}
    \bigvee_{j:w_j^{(d)}=1}z_j=1 &\iff \exists j\; w_j^{(d)}=1 \land z_j=1\\
    &\iff \max\left\{w^{(d)}_j z_j; j=1,...,J\right\}=1
\end{align}

\section{Loss function and training}
\label{appendix:loss}
Our loss function has a mean-square error component together with components that promote sparsity of the obtained rule model. The overall loss function is:
\begin{equation}
    \mathcal{L}_{mse} + \lambda_{and} \|\mathbf{W}\|_1 + \lambda_{or} \|\mathbf{w}^{(d)}\|_1 + \lambda_{p} \|\mathbf{W}^{(p)}\|_1 
    \label{eq:loss}
\end{equation}
where $\|\cdot\|_1$  is the 1-norm. Increasing $\lambda_{p}$ reduces the sparity of the learned literals, increasing $\lambda_{and}$ shortens conjunctions, and increasing $\lambda_{or}$ shortens the disjunction. For simplicity, we choose a single regularization penalty for all the different layers $\lambda = \lambda_{and} = \lambda_{or} = \lambda_{p}$. Unless stated otherwise, we use $\lambda = 10^{-2}$ for training on synthetic datasets (with known ground truth) and a more conservative value of $\lambda=10^{-3}$ for the UCI ML datasets \cite{Dua:2019}.  

We use the Adam optimizer from the PyTorch ecosystem with default learning rate and parameters to optimize Eq. \ref{eq:loss}. In order to avoid local minima throughout the training routine, we randomly re-initialize the weights of the neural network every $10^4$ epochs and run the network for $2.5 \times 10^5$ epochs in total. The training is done with a fixed batch size of 100. We select the epoch with the lowest loss function on the training set and report the prediction accuracy at that stage for the test-set (train/test split of 80/20).

As mentioned in the section on the literal learning module, to ensure smooth training as well as reaching binary output of the literal learning layer, we propose a cooling schedule of the sigmoid activation function with high temperature in the beginning allowing larger gradients to exist across a wider range of the input so that training can occur. As the temperature approaches zero, the sigmoid is progressively scaled and approaches the Heaviside function in the limit, so that we achieve strict binarization in the end. The temperature is cooled down with a factor $\gamma = 0.995$ at every subsequent epoch for all the examples presented in the paper.  

The number of output nodes of the literal learning module $m = 10$ while the number of nodes in the rule learning module is $J=25$. Increasing or decreasing these numbers would impact the expressivity of the network but we empirically observed that varying this number in the range 10-50 did not show a significant sensitivity to these values.

\section{Sparsity of the DNF} 
\label{appendix:sparsity}
In the examples presented in Table 1, the number of conjunctions in the DNF is always smaller than the number of possible conditions in the hidden layer of the rule learning module. As described in the Appendix C, the loss function of R2N contains sparsity regulation on the weights of the literal learning, AND and OR-layer, controlling the sparsity of the resultant DNF. To assess the role of the sparsity promoting parameter, $\lambda$, we trained the R2N for values of $\lambda$ in the range $[10^{-5},1]$, both for Ex. 4 and 5 from Table \ref{tab:1}. As shown in Fig. \ref{fig:3} (red), the number of conjunctions in the DNF decreases upon increasing $\lambda$. The accuracy however (black), remains near optimal for values inferior to $\lambda < 10^{-1}$, suggesting that the most compact DNF that has the smallest compromise on accuracy can be obtained for values of $\lambda \approx 10^{-2}$ or $10^{-3}$. At larger values of $\lambda$, the strength of the sparsity constraint will result in DNFs that are shorter than the ground-truth, resulting in approximations of the hyperplanes that are sparser, and compromise on the predicition accuracy, e.g. for Ex. 4, $ 77\%$ accuracy is obtained with a DNF containing a single conjunction $(-x_0 -0.7 x_1 - 0.4 x_3 + 0.3 x_4 > -0.7)$, while the ground-truth DNF contains three conjunctions. Controlling the value of $\lambda$ provides the user an opportunity to balance the prediction accuracy with the sparsity of the DNF.

\section{DNFs obtained with R2N}
\label{appendix:rulesets}
In this section we show the DNFs obtained with R2N and discuss how they are obtained. The DNF associated to the learned rule model can be extracted directly from the binary weight matrix $\mathbf{W}$ and $\mathbf{w}^{(d)}$, while the values of the learned literals are extracted from the weight matrix $\mathbf{W}^{(p)}$. Since the values of $\mathbf{W}^{(p)}$ are constrained with a sparsity penalty (see Appendix \ref{appendix:loss}), most values are small, yet not strictly zero. We therefore threshold all the values that have a normalized coefficient that is smaller than $2.5 \%$ of the magnitude of the largest coefficient in a predicate. For instance, for example 5 in Table \ref{tab:1}, the unprocessed output of R2N is given by, 
\begin{equation}
\small
\begin{split}
& [(-0.7x_0-0.7x_1-x_2-0.2x_3-0.8x_4 > -1173) \land (x_4 < 0.2) \land (x_0-0.6x_1 > -0.1)]  \\ 
\lor & [(x_4 < 0.2) \land (-x_0-0.4x_1-0.1x_2-0.2x_4 > 704)] \\
\lor & [x_0 < 0.2] \\
\lor & [(0.5x_0-0.2x_1-x_2-0.4x_3+0.1x_4 > -540) \land (-x_0-0.4x_1-0.1x_2-0.2x_4 > 704)] \\
\lor
& [(x_4 < 0.2) \land (0.3x_0+0.1x_2-x_3-0.3x_4 > 3050)] \\
\lor
&  [0.4x_1+x_3 > 1.0]
\end{split}
\end{equation}
This can be simplified by identifying and removing all the constant false ($\bot$) or true ($\top$) predicates (always true or false), e.g.,  $0.5x_0-0.2x_1-x2-0.4x_3+0.1x_4 > -540 \rightarrow \top$, into,
\begin{equation}
\small
\begin{split}
& [(x_4 < 0.2) \land (x_0-0.6x_1 > -0.1)]  \\ 
\lor & [x_0 < 0.2] \\
\lor & [0.4x_1+x_3 > 1.0]. 
\end{split}
\end{equation}

The simplified DNFs we obtain for the other examples in Table 1 are: 
\newline
For ex. 1: 
\begin{equation}
\begin{split}
& [(x_0 > 0.26)] \\ 
 \lor & [(x_1 < 0.49)].
\end{split}
\end{equation}
For ex. 2: 
\begin{equation}
\begin{split}
&[(x_0-0.41x_1 > 0) \land (x_0-0.53x_1 > 0)] \\
\lor & [(x_4 < 0.25)].
\end{split}
\end{equation}
For ex. 3: 
\begin{equation}
\begin{split}
&[(x_0+0.49x_1 > 0.41) \land (x_4 < 0.23)] \\
\lor & [(x_0+0.49x_1 > 0.41) \land (x_0+0.50x_1 > 0.51)] \\
\lor & [(x_4 < 0.25)].
\end{split}
\end{equation}
Finally for ex. 4: 
\begin{equation}
\begin{split}
(x1 < 0.49) \land (-x_3+x_4 >= 0) \\
\lor (x_0 < -0.20) \\
\lor [(x_0-0.1x_1-0.2x_2-0.4x_3-0.2x_4 > -2.2) \land (x_0 < 0.2) \land (-x_3+x_4 > 0)]  \\
\lor [ (x_0-0.1x_1-0.2x_2-0.4x_3-0.2x_4 > -2.1) \land (x_4-x_3 > 0) \land (0.1x_0+x_1+x_2-0.1x_3+ > 0.46)
]
\end{split}
\end{equation}

\section{Baselines}
\label{appendix:baselines}

This section describes the configuration settings for the three baselines: DR-Net \cite{qiao2020learning}, RIPPER \cite{cohen1995fast}, and BRCG \cite{dash2018boolean} used in our experiments. For all cases, we used a common training set of size 8000 and evaluated on a common test set of size 2000, and used the equi-quantile feature binarizer included in the BRCG repository\footnote{\url{https://github.com/Trusted-AI/AIX360/blob/master/aix360/algorithms/rbm/features.py}} with 40 bins.

For BRCG, we use the publicly available implementation from the AI Explainability 360 repository\footnote{\url{https://github.com/Trusted-AI/AIX360}}. The algorithm uses two parameters $\lambda_1$ and $\lambda_2$ to penalize the learning of more and longer clauses respectively. We set $\lambda_1 = \lambda_2 = 0.001$, which is the default value recommended in the implementation. We also use the default values for the maximum number of iteration (100), maximum number of columns generated per iteration (10), and maximum degree (10) and width (5) parameters for the beam search.

For DR-Net we use the implementation by the authors\footnote{\url{https://github.com/Joeyonng/decision-rules-network}} with the default parameters except for the number of training epochs, which was increased from 2000 to 10000 according to the authors' recommendation to learn sparse rules. For RIPPER, we used the wittgenstein implementation\footnote{\url{https://github.com/imoscovitz/wittgenstein/tree/master/wittgenstein}}. But instead of using its internal binarizer, we binarized it upfront using the aforementioned feature binarizer.

For evaluating the predictive accuracy for the synthetic examples in Table 1 in the main text we also report the values obtained from a MLP (Multi Layer Perceptron). We used a simple two layer architecture, with 10 neurons in the hidden layer, trained for $10^{4}$ epochs with default optimizer parameters (Adam optimizer).  

\begin{figure}[t]
    \centering
    \includegraphics[width=0.45\textwidth]{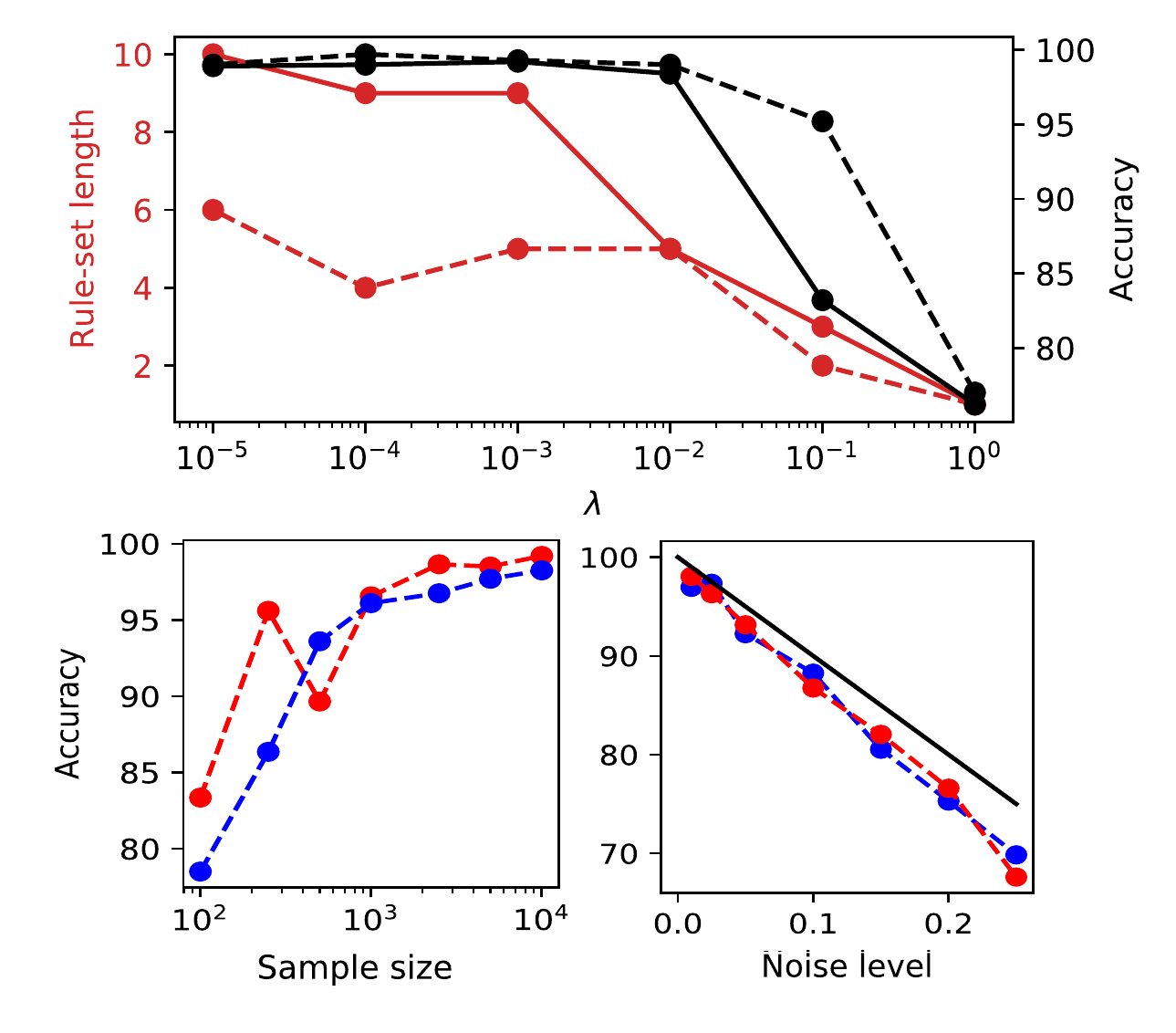}
    \caption{(top) Accuracy on the test-set and number of conjunctions in the DNF, $n_r$, as function of  $\lambda=\lambda_{and}=\lambda_{or}=\lambda_{p}$ for Ex. 4 (dashed) and Ex. 5 (solid) from Table 1. Accuracy on the test set as function of (left) the size of the training set and (right) the noise level for (red) Ex. 4 and (blue) Ex. 5 from Table 1.}
    \label{fig:3}
\end{figure}

\section{Robustness w.r.t. noise and sample size}
\label{app:noisesample}
Data from which rule models are learned are oftentimes noisy and sparse. To assess the robustness of our approach w.r.t. number of training examples, we plot the prediction accuracy on the test set for Ex. 4 and 5 of Table \ref{tab:1} as function of the size of the training set (Fig. \ref{fig:3}) and find that, even for a training set size of $O(10^3)$ we obtain a good accuracy ($> 0.95 \%$). This shows that our method is not overly sensitive to the size of the training dataset, a common problem with neural-network based models. 

Alternatively, when we add white noise to the data provided to the algorithm (randomly flipping a fraction of the labels in the dataset) we find that for both examples the accuracy of the test-set approaches near-optimal accuracy up to noise levels of $10\%$ (Fig. \ref{fig:3}: The solid line indicates optimal performance, note that since the data is noisy, this line is the upper limit for the performance). 

\section{Interoperability with other rule learning algorithms}
\label{app:interop}
The relational literals that are learned by R2N can also serve in other applications as part of the domain vocabulary (or ontology); for instance as input literals for other rule learning algorithms, ensuring interoperability with existing rule learning workflows. To showcase how this can augment algorithms like RIPPER, we used the \emph{learned} literals from R2N as input literals for RIPPER and found that for all synthetic examples in Table \ref{tab:1}, the number of rules learned is considerably less as compared to RIPPER with univariate value comparisons. As we display in Table \ref{tab:custom_predicates}, the learned number of rules decreases and becomes comparable with that of R2N, along with an improvement in accuracy for 4 of the 5 cases (not shown in the table), demonstrating that the learned literals are not only specific to R2N but can also benefit other rule learning algorithms. 

\begin{table}[t]
    \small
    \centering
    \begin{tabular}{|c|c|c|c|c|c|}
     \hline
     Example.& 1 & 2& 3& 4& 5 \\
              \hline

    $n_r$ with quantile binning & 5 & 22 & 15 & 23 & 15 \\

         \hline
    $n_r$ with learned literals & 3 & 3 & 2 & 5 & 3 \\
     \hline
    \end{tabular}
    \caption{Number of conjunctions obtained with RIPPER by using quantile binning and the \emph{learned} literals obtained from R2N for the 5 synthetic examples presented in Table 1}
    \label{tab:custom_predicates}
\end{table}

\end{document}